# Interpreting CNN Predictions using Conditional Generative Adversarial Networks

R T Akash Guna[a], Raul Benitez[b,c,*] and O K Sikha[a,b,*]

[a]*Department of Computer Science and Engineering, Amrita School of Computing, Coimbatore, Amrita Vishwa Vidyapeetham, India,*
[b]*Automatic Control Department, Universitat Politècnica de Catalunya (UPC-BarcelonaTECH), 08034 Barcelona, Spain,*
[c]*Institut de Recerca Sant Joan de Déu (IRSJD), 08950 Barcelona, Spain,*

ABSTRACT

We propose a novel method that trains a conditional Generative Adversarial Network (GAN) to generate visual interpretations of a Convolutional Neural Network (CNN). To comprehend a CNN, the GAN is trained with information on how the CNN processes an image when making predictions. Supplying that information has two main challenges, namely how to represent this information in a form feedable to the GANs and how to effectively feed the representation to the GAN. To address these issues, we developed a suitable representation of CNN architectures by cumulatively averaging intermediate interpretation maps. We also propose two alternative approaches to feed the representations to the GAN and to choose an effective training strategy. Our approach learned the general aspects of CNNs and was agnostic to datasets and CNN architectures. The study includes both qualitative and quantitative evaluations and compares the proposed GANs with state-of-the-art approaches. We found that the initial layers of CNNs and the final layers are equally crucial for interpreting CNNs upon interpreting the proposed GAN. We believe training a GAN to interpret CNNs would open doors for improved interpretations by leveraging fast-paced deep learning advancements. The code used for experimentation is publicly available at
https://github.com/Akash-guna/Explain-CNN-With-GANS

## 1. Introduction

Neural networks have become the backbone of most artificial intelligence advancements in recent times. Despite their success, how a neural network makes its predictions is not directly understandable by humans. A Convolutional Neural Network (CNN) is a neural network designed to process pixel data [24] using trainable convolutional filters and has become the most popular approach to solve many image-related problems[43] such as image classification[18, 20], recognition [36, 33] and segmentation[17, 31]. Explainable AI is a domain aimed at resolving this issue[12]. It is essential to interpret CNNs to find whether they primarily use expected regions to make predictions. Zhou *et al*.[52] used a global average pooling layer to construct class activation maps (CAM) to interpret CNNs. Variations of CAM [34, 5, 45] are successful in interpreting CNNs, but mathematical approximations often suffer from errors.

We develop this research based on recent advancements in deep learning interpretability [34] and conditional GANs [47]. GANs are used for various image translation [21, 25, 48] tasks therefore, it is feasible for a GAN, when fed with an input image, to generate interpretations on how a CNN made its predictions for the image. We investigate whether the interpretations produced by the GAN are better than existing methods to interpret CNNs. Additionally, we interpret the interpretations of the proposed GAN to identify important CNN layers for making predictions. To the best of our knowledge, we are the first to experiment GANs to interpret CNN predictions.

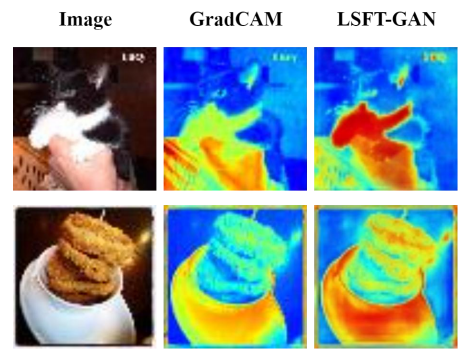

**Figure 1:** Sample Results of the LSFT-GAN interpreting CNNs compared with Grad-CAMs

A GAN to interpret CNNs should learn to understand CNNs regardless of datasets and CNN architectures. Further, we believe that to interpret CNNs, a GAN needs to be conditioned with explicit knowledge on how a CNN identifies patterns from an image to make its predictions. This requirement causes two primary design issues, namely how to represent the working of CNNs as conditions and how to feed the conditions to the GANs. We design and train an architecture addressing these issues by representing the required knowledge as a series of cumulative averages of intermediate Grad-CAM [34] maps (CGMA). We introduce two methods for conditioning CGMAs into the GAN, representing all the CGMAs as a single global condition that conditions the GAN at regular intervals and conditioning the GAN by feeding CGMAs progressively (Local). The conditions are transformed spatially and feature-wise using a Spatial Feature Transform Layer (SFT) and then added to the GAN. Therefore, we denote these two methods as Global SFT (GSFT) and Local SFT (LSFT) GANs.

*Corresponding author
cb.en.u4cse19302@cb.students.amrita.edu (R.T.A. Guna);
raul.benitez@upc.edu (R. Benitez); sikhakrishnanunni@gmail.com (O.K. Sikha)

During experimentation, it was found that both SFT-GANs captured additional regions of interest not recognized by Grad-CAM, the state-of-the-art interpretation model. These regions were synonymous with regions identified by saliency-based interpretation methods [37]. We also identified that LSFT-GAN is more robust and provides more general descriptions of a CNN than GSFT-GAN. The LSFT-GAN performed better on unseen datasets and CNN architectures compared to Grad-CAM. During the interpretation of LSFT-GAN interpretations, we found that the LSFT-GAN relied equally on initial and final CNN layers for CNNs. We summarize our contributions as follows:

1. We introduce a conditional GAN that could understand how CNNs work in general.
2. We introduce CGMA, a method to represent explicit knowledge of how CNN operates in a form feedable to the GAN.
3. We introduce LSFT and GSFT GANs and identify that LSFT-GAN is more effective to interpret CNNs through experiments. It is also found that the LSFT-GAN performs better than existing methods on unseen data and architectures, as shown in Figure 1.
4. We interpret the LSFT-GAN to understand how it interprets CNNs.
5. We conduct an extensive set of experiments to investigate the explainability and the accuracy of the proposed model both quantitatively and qualitatively on classification and image captioning tasks. A new evaluation metric, "Robustness towards image degradation," was introduced to analyze the interpretation capacity of the proposed method in the presence noise.

The ordering of the sections is as follows: the background of this paper is discussed in Section 2. The proposed GAN architecture and design choices are detailed in Section 3. In Section 3.1 we discuss the proposed conditions that represent knowledge of a CNN's operations. Section 3.2 elaborates the conditioning strategies and general architecture of the GAN. Section 4 discusses the training strategies followed and the results of quantitative and qualitative experimentation. We discuss the datasets used and training strategies adopted in Section 4.1. In Sections X and 4.2 we perform experiments to assess the SFT-GANs both qualitatively and quantitatively. We evaluate the robustness of the GANs (both LSFT and GSFT) towards degradation by the addition of noise in Section 4.4. We then interpret the trained LSFT-GAN to investigate how the LSFT-GAN interpreted CNNs, these experiments are presented in Section 5. The limitations of the proposed GAN are presented in Section 7. We then conclude the paper by chronologically summarizing the paper in Section 8.

## 2. Background

### 2.1. CNN Explainability

Recently, several methods for interpreting CNNs have been proposed [51, 42, 52, 34, 5, 45]. Deconvolution was introduced by Zeiler *et al.*[51], which involved layer activation from higher layers flowing into the input image. Deconvolution was then improved with guided backpropagation [42] to produce maps representing each CNN neuron. Bolei *et al.*[52] proposed class activation maps (CAM), which uses a global average Pooling layer to interpret the underlying CNN. Selvaraju *et al.*generalized CAM by introducing Grad-CAM [34], a method that combines the concepts of deconvolution and guided backpropagation to eliminate the need for the global averaging layer as in class activation maps (CAM). Aditya *et al.*[5] proposed Grad-CAM++, which improved CAM in cases where multiple objects appeared in a single image and provided more robust interpretation maps. By performing a forward pass to obtain the weight of each activation map for a target class, ScoreCAM [45] removed the dependency of gradients. In this study, we use intermediate layer Grad-CAMs to represent CNN as conditions in this paper, which is discussed in Section 3.1.

### 2.2. Generative Adversarial Networks

A Generative Adversarial Network (GAN) [16] is an unsupervised neural network used for generating unstructured data, i.e. images. It is composed of two parts, namely a generator and a discriminator. The generator uses latent space to build a synthetic image, and the discriminator is trained to recognize the difference between a real and a synthetic image. The generator tries to fool the discriminator, while the discriminator trains to distinguish between synthetic and real images.

### 2.3. Conditional GANs

A regular (unconditional) GAN generates images of random classes, but the GAN architecture can be conditioned to generate images of a specific class, resulting in what is known as a conditional GAN (cGAN) [28]. Isola *et al.*[21] proposed conditional GANs to solve cross-domain image-to-image translation problems, where a conditional GAN learned the mapping from the input to the output domain. The study also proposed a loss function to train this mapping. In addition, conditional GANs have been successfully applied in problems such as image super-resolution [25], image blending [48] and image restoration [46]. Wang *et al.*[47] proved that the performance of image-conditioned GANs could be improved by using additional conditions that provide relevant characteristics. In our proposed approach, a suitable representation of the CNN processing workflow is used as a condition to train the conditional GAN architecture.

### 2.4. Image-to-image translation using GANs

An image domain is a set of images presenting similar acquisition conditions such as illumination, signal-to-noise, or spatial resolution [23]. Image-to-image translation is the process of transforming an image from one domain to another. Isola *et al.*introduced Pix2Pix-GAN [21] for cross-domain image translation. CycleGAN [53] removed the dependency of paired images for translation by cyclically training two generators. StarGAN [7] reduced the need for additional generators by incorporating the domain to be transformed as a condition of the generator. Numerous tasks such as virtual try-on [19, 14], image cleaning [50, 32], segmentation [6, 30], super-resolution [25, 2] and restoration [46] have emerged since the introduction of image translation GANs. This paper uses a GAN architecture to translate an input image into an interpretation heatmap. The feature maps learned by different layers of the CNN model are provided as conditions to the GAN to aid the translation.

## 3. Proposed Methodology

This section introduces the proposed GAN model for interpreting CNN-based predictions. Two fundamental design issues must be addressed when developing GANs to interpret a CNN:

1. How to represent the operation of different CNN architectures as conditions?
2. How to effectively feed the conditions to the GAN?

The following sections will discuss our approach to address these issues and introduce our model construction pipeline.

### 3.1. Representation of CNNs

As discussed in Section 1, to interpret how a CNN makes predictions, GANs should be conditioned with information about how a CNN operates images. To replicate how a CNN learns, we investigate the use of gradient-based interpretation maps as the conditions for the GAN. Existing gradient-based techniques [34, 5] interpret predictions using gradients from the last convolutional layer of a CNN architecture. Although the gradients from the last convolutional layer capture the relevant features of all preceding layers, it is prone to information loss since the final heat map is an approximation from the final convolutional layer. As a consequence, a single Grad-CAM does not completely represent all the details of a CNN architecture.

Consider $Conv_i$ be the i-th convolutional layer in a CNN with $N$ convolutional layers. Grad-CAM from layer $Conv_{i+1}$ would be able to represent the layer $Conv_i$ better than $Conv_j$ where $i + 1 < j \leq N$, caused by an effect known as the vanishing gradient problem [4]. Therefore, the best scenario is to condition the GAN using Grad-CAMs from all convolutional layers but, it is not feasible due to a varied number of convolutional layers in different CNN architectures. In CNNs with a high number of convolutional layers selecting a sparse number of layers would not result in a fitting approximation hence, it is required to fit all preceding layer information into each interpretation map. We solve this by representing CNNs as a series of cumulative Grad-CAM averages.

#### 3.1.1. Cumulative Grad-CAM Averages (CGMA)

Since the Grad-CAM of a single layer is not sufficient to represent the complete model, we compute *Intermediate Grad-CAMs* for all convolutional layers of the CNN architecture. Intermediate Grad-CAMs are computed for each $Conv_i$ where $Conv_i$ is the i-th convolutional layer in a CNN with N layers. A fixed number $\eta$ of Grad-CAMs are selected from the N layers using Equation 1 to minimize the variable number of layers problem (in case of different CNN architecture). The selection of Grad-CAMs ensures that the Grad-CAMs of the first and last layers of CNN are included in the CGMA representation. In Section 4 we use a value of $\eta = 8$ for the empirical evaluation.

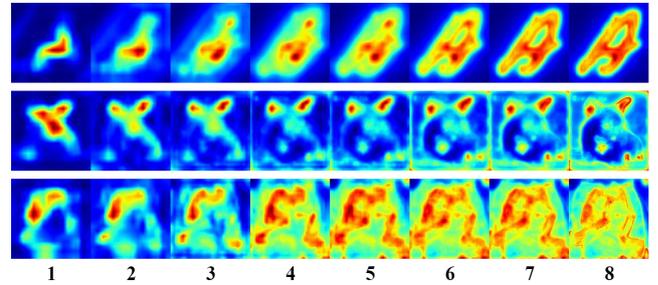

**Figure 2:** CGMAs of 3 images each from EMINST [9], Animals-10 [3] and Cats Vs Dogs [13] datasets were obtained from a VGG16 network with 12 convolutional layers. 8 CGMAs are selected using Equation 1. Each column has 8 CGMAS starting from the initial layers to the final layers of the CNN

$$\text{Positions}[K] = \lceil \lceil \text{skip} \times K \rceil \times \frac{N}{100} \rceil \quad (1)$$

where K is in the range of $(0, \eta-1)$, N is the number of layers in CNN, and skip is given by Equation 2

$$\text{skip} = \frac{100}{\eta - 1} \quad (2)$$

The variable number of layers problem was addressed by selecting $\eta$ Grad-CAMs, but the vanishing gradients' problem persisted when $\eta \ll N$.

$$\text{CGM}[i] = \sum_{j=1}^{i} \text{Grad-CAM}[j] \quad (3)$$

The CGM from the selected layers is then averaged to form CGMAs. The final formulation of CGMA is shown in Equation 4

$$\text{CGMA}[i] = \frac{\text{CGM}[i]}{\text{Positions}[i]} \quad (4)$$

CGMAs obtained using Equation 4 for a character *"A"* from the EMINST dataset [9], a mouse from Animals 10 dataset and a dog from Cats vs Dog dataset [13] each classified using a VGG16 architecture are shown in Figure 2. In the study, a total of 8 CGMAs among the 12 convolutional layers of the VGG16 architecture were selected using Equation 1.

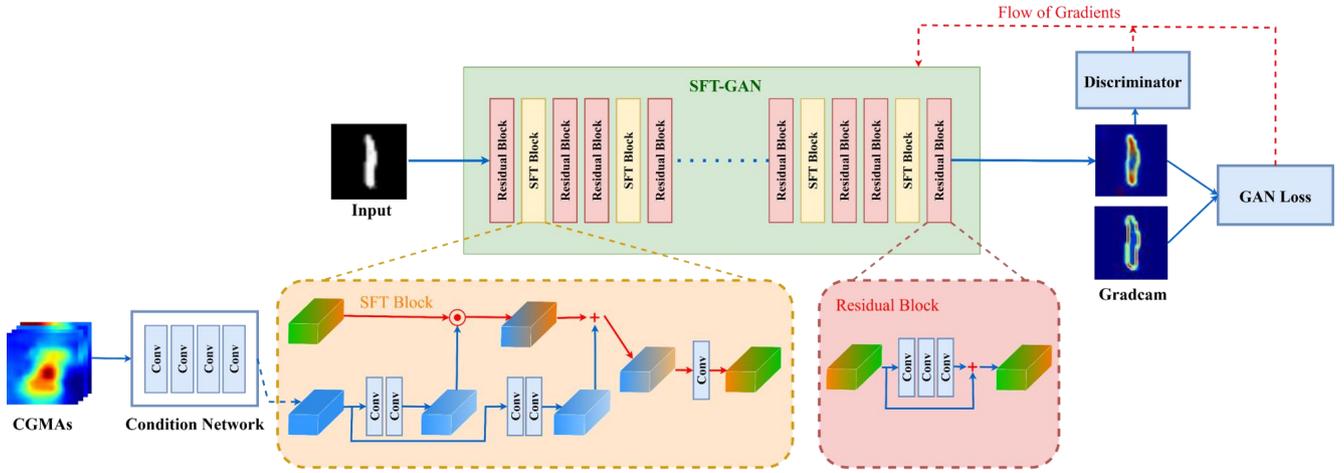

**Figure 3:** Model Architecture. The input image and the CGMAs for the dataset-model pair are passed into the GAN to generate interpretations. CGMAs are passed into SFT Layers through a Condition Network to reduce the dimensionality of the vector. The proposed GAN consists of SFT and residual blocks. The number of SFT blocks is  . Each SFT block contains an SFT Layer followed by a convolutional layer. Each residual block contains 3 convolutional layers, with a skip connection at the final layer. A discriminator Loss and a GAN Loss are used for updating the weights of the GAN during training

### 3.2. Building the GAN

This section describes the design of the proposed SFT-GAN architecture for interpreting CNNs. The primary question is how to condition the GAN using CGMAs. We tested two approaches for supplying CGMAs as conditions to the GAN. The first method (Local-SFT, LSFT) uses local conditioning of the GAN with several CGMAs whereas the second method (Global-SFT, GSFT) uses single global conditioning. This section describes the architecture of both of these models, which will be then compared in Section 4.

#### 3.2.1. Spatial Feature Transform (SFT)

Spatial Feature Transform (SFT) layer was introduced by Wang *et al*.[47] for image super-resolution. We use an SFT layer to learn a mapping function     that produces a modulation parameter pair $(\alpha, \beta)$ based on the CGMAs as described in Equation 5. The learned modulation parameters influence the output by applying a spatial affine transformation to the network's intermediate feature maps. As we use SFT layers in our proposed architecture, we hereon refer to the proposed GAN as SFT-GAN.

$$: \text{CGMA} \mapsto (\alpha, \beta) \quad (5)$$

The SFT transformation is then carried out by scaling and shifting the individual feature maps using the obtained modulation parameter pair ( , ).

$$\text{SFT}(Feat \mid \alpha, \beta) = \alpha \odot Feat + \beta \quad (6)$$

Where *Feat* is the feature map and $\odot$ is the Hadamard product of the operands (element-wise multiplication). The dimension of the feature map is the same as the dimension of the modulation parameters $(\alpha, \beta)$, so the SFT layer performs spatial transformation along with feature manipulation. Spatially altering the conditions alongside feature manipulation helps the GAN to localize regions of interest from the conditions faster. Figure 3 shows the generic working of SFT-GANs. We trained two SFT-GANs architectures that differed in terms of their conditioning strategy: a globally conditioned GAN (GSFT-GAN) and a locally conditioned GAN (LSFT-GAN).

#### 3.2.2. GSFT-GAN

The GSFT-GAN uses global conditioning to train the SF-GAN. All the conditions (CGMAs) generated from the convolutional layers of CNN are used together to form a global condition which is then fed to all SFT layers of the GAN as proposed in existing GANs with SFT layers [46, 47]. Figure 4 shows the architecture of GSFT-GAN.

#### 3.2.3. LSFT-GAN

A CNN learns high-level image features in the initial layers and then progresses to learn complex features as we go deeper into the layered architecture [24]. Because CNN learns incrementally, the initial layers are critical in extracting these complex features. The main drawback of the GSFT-GAN approach is that since it employs global conditioning feature evolution across the CNN architecture, the incremental learning of CNN may get overlooked. Therefore, we explore a method for locally conditioning the GAN to preserve the incremental spatial information from all CNN layers. Each CGMA is passed into the condition network separately, resulting in   conditions, which are then transformed using the SFT layers. The order in which the conditions are passed into SFT layers corresponds to the flow of the CNN architecture under consideration. For instance, if     and     are convolutional layers of a CNN, CGMA[ ] and CGMA[ ] are their respective CGMAs, and SFT   and SFT   are SFT layers of LSFT-GAN, assuming that   <    and (   <   ) <=   . The condition prior CGMA   will be passed into SFT   and CGMA   will be passed into SFT   as shown in Figure 4.

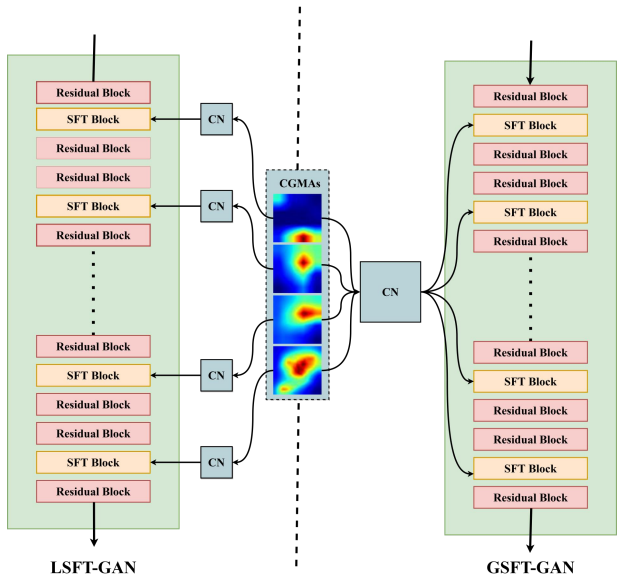

**Figure 4:** LSFT-GAN and GSFT-GAN Architectures. Both the architectures differ by how the CGMAs are conditioned. In LSFT, the SFT-GAN is conditioned locally by conditions where each CGMA is a condition, whereas the GSFT-GAN is conditioned using a global condition containing all CGMAs. Here *CN* stands for condition network

## 4. Experimental validation of the approach

In this section, we conduct experiments with different datasets to evaluate the effectiveness of the proposed SFT-GAN interpretations. Section 4.1 describes the datasets used for the experiments and training strategies used to train the SFT-GANs. In Section 4.3 we qualitatively evaluate our approach by comparing the interpretation heatmaps generated by SFT-GANs with state-of-the-art interpretability approaches. Then we quantitatively compare LSFT, GSFT, and Grad-CAM in Section 4.2. Visualization results under different signal-to-noise conditions are provided in Section 4.4.

### 4.1. Datasets and implementation

We adopted a training strategy that uses multiple datasets and classification models to train the SFT-GANs to produce general interpretations of CNNs. Since LSFT-GANs and GSFT-GANs differ only by their architecture, their training process remains the same.

#### 4.1.1. Datasets used for training

We explored 2 different strategies to train the SFT-GANs. The aim was to assess which strategy provides better generalization and to assess whether we need specialized SFT-GANs to interpret crucial tasks. Firstly, we trained the GANs using 3 image datasets of similar semantics [11, 9, 29] with images from different alphabets. Secondly, we trained the SFT-GANs using 3 image datasets of dissimilar semantics [22, 39, 3] presenting a larger variety of recognition classes and semantic categories.

**Similar Semantics Strategy** Three different datasets were employed for the experimentation: MNIST [11] a dataset of handwritten digits, EMNIST [9] an extension of MNIST for English alphabets and Chinese MNIST [29] dataset, which consists of handwritten Chinese characters. We used 90% of the data for training and 10% for testing CNNs. These datasets where similar symbol databases, hence shared semantics such as symbol position, background and colors.

**Dissimilar Semantics Strategy** We chose CIFAR-10 [22], Animal-10 [3] and Food-11 [39] datasets for experimentation. CIFAR-10 has 10 classes with images of the size 32 × 32 pixels closely cropped to the objects. CIFAR-10 has 5000 images for training and 1000 images for testing. Animal-10 contains 10 categories of animals collected from the internet. These images had a considerable portion of the background. Each image had varied dimensions. The images of the Food-11 database had 11 categories of food with images with background objects such as plates, cutlery, and glasses. The dimensions of the images were scaled to 512 if the original dimension was greater than 512 pixels. From the descriptions of the three datasets it is clear these datasets vary in different semantics such as objects, background, and quality hence, were chosen for training the SFT-GAN using strategy-2.

**CNN architectures for training SFT-GANs** In order to describe how our approach works, we used 3 different CNN architectures for training the SFT-GAN namely VGG16 [38], ResNet50 [18] and MobileNet [38]. VGG16 has a total of 16 layers. VGG16 has 13 convolutional layers in series and 3 dense layers. Max-pooling layers are used for reducing the dimensions of the tensors and to avoid overfitting. ResNet50 has 48 convolutional layers, 1 max-pooling layer, and 1 average-pooling layer. The convolutional layer of the ResNet included skip connections to resolve the vanishing gradient problem [4]. MobileNet is a CNN architecture designed to perform on lighter systems. MobileNet uses depthwise separable and pointwise separable convolutions that reduce the number of computations required as compared to a traditional convolutional layer. These 3 CNN architectures were chosen as they are structured differently and thus help the trained SFT-GAN to generalize between multiple architectures.

**Training CNNs** To train the cGAN, we previously trained CNN models for classifying 3 datasets using 3 architectures per training strategy, thus resulting in a total of 9 CNNs trained for each training strategy. For the first strategy, each of the nine models had a testing accuracy greater than 95%. Models trained on Chinese MNIST Dataset [29] had an average testing accuracy of 99%. For the second strategy, each CNN had an average testing accuracy over 90% . Models trained on Animals-10 dataset had an average test accuracy of 95%. For both strategies, the CNNs were transfer learned from ImageNet weights [10]. Each model was trained for 20 epochs. Adam optimization algorithm was used to optimize the gradients for ResNet and MobileNet architectures, while SGD performed better for VGG16. The

| Dataset | Model | GSFT | | | | LSFT | | | |
|---|---|---|---|---|---|---|---|---|---|
| | | TP | FP | TN | FN | TP | FP | TN | FN |
| MNIST | VGG16 | **0.8590** | 0.0100 | 0.0030 | 0.9064 | 0.5972 | **0.0093** | 0.0026 | **0.9312** |
| | MobileNet | **0.9151** | 0.0090 | 0.0020 | 0.9130 | 0.8797 | **0.0038** | **0.0005** | **0.9421** |
| | ResNet | **0.9532** | **0.0040** | 0.0030 | 0.9234 | 0.9250 | 0.0057 | **0.0003** | **0.9473** |
| EMINST | VGG16 | **0.9470** | 0.0047 | **0.0002** | 0.9090 | 0.8952 | **0.0026** | 0.0003 | **0.9318** |
| | MobileNet | **0.8905** | 0.0008 | 0.0001 | 0.9228 | 0.8622 | **0.0004** | 0.0001 | **0.9512** |
| | ResNet | **0.8828** | 0.0026 | **0.0001** | 0.9080 | 0.8607 | **0.0001** | 0.0002 | **0.9466** |
| Chinese | VGG16 | 0.5041 | **0.0050** | 0.0018 | 0.9577 | **0.6680** | 0.0064 | 0.0018 | **0.9608** |
| | MobileNet | 0.5350 | **0.4640** | 0.0016 | **0.2706** | **0.8862** | 0.5338 | **0.0007** | 0.2248 |
| | ResNet | 0.5410 | **0.0050** | 0.0010 | 0.9638 | **0.7448** | 0.0074 | **0.0008** | **0.9644** |

**Table 1**
Quantitative analysis of interpretability maps produced by GSFT-GAN and LSFT-GAN. Here the best value is highlighted by different colors. The color is RED when higher value is better than lower value and BLUE when vice-versa learning rate used was $10^{-3}$. None of the trained models were overfitted, as the testing accuracy of all models is relative to their training accuracy.

**Training SFT-GANs** When training the SFT-GANs, the data was split by classes. As a result, the model was unaware of the testing data classes during SFT-GANs training. This method of splitting was adopted to estimate the generalization capacity for interpreting CNNs. All nine dataset-model combinations were used for training the SFT-GANs. We calculated CGMAs for each element of the pair as in Equation 4. We used Grad-CAM heatmaps [34] as target images for the generator. The discriminator's objective is to differentiate between a Grad-CAM heatmap and the generated heatmaps. We used the mean squared error as the generator's objective function and binary cross entropy as the discriminator's objective function.

### 4.2. Quantitative Assessment
#### 4.2.1. Comparison of LSFT-GAN and GSFT-GAN

In this section, we compared LSFT-GAN and GSFT-GAN based on their ability to localize objects and differentiate between foreground and background in the images. We calculate the number of true positive pixels (foreground as foreground), true negative (background as background), false positive (foreground as background), and false negative (foreground as foreground) to evaluate localization, and the results are tabulated in Table 1. We performed this experiment only with strategy-1 due to the unavailability of ground truths for the datasets used in strategy-2. LSFT-GAN is more precise and trustworthy than GSFT-GAN since it has a greater FN and FP across all three datasets and all CNNs considered for the study (bolded in Table 1). GSFT-GAN showed better localization of foreground on MNIST [11] and EMINST dataset [9] (high TP in Table 1), whereas LSFT-GAN provided good performance at identifying background regions (better TN in Table 1). Although GSFT-GAN showed better foreground localization, it is still ambiguous whether these are relevant regions. In Section 4.2.3, we identify if the regions are relevant.

| Dataset | Models | Grad-CAM | GSFT | LSFT |
|---|---|---|---|---|
| MNIST | VGG16 | 0.4557 | **0.6531** | 0.5543 |
| | MobileNet | 0.6411 | **0.6664** | 0.5579 |
| | ResNet | 0.6799 | 0.6854 | **0.7040** |
| EMINST | VGG16 | **0.6772** | 0.6573 | 0.6757 |
| | MobileNet | 0.6361 | **0.6585** | 0.6440 |
| | ResNet | 0.6372 | **0.6607** | 0.6451 |
| Chinese | VGG16 | 0.5093 | 0.4998 | **0.5926** |
| | MobileNet | **0.7110** | 0.4870 | 0.6510 |
| | ResNet | **0.9988** | 0.7934 | 0.9042 |
| **Average** | | 0.6577 | 0.6401 | **0.6588** |

**Table 2**
Comparison of LSFT-GAN and GSFT-GAN with Grad-CAM using Sørensen–Dice coefficient for each dataset-model pair. The largest value among a row is marked in RED. The greatest dice coefficient is highlighted in BLUE given dice coefficient of Grad-CAM was marked RED

#### 4.2.2. Comparison with Grad-CAM

This section compares Grad-CAM and proposed SFT-GANs quantitatively using Sørensen–Dice similarity coefficient [41]. The dice similarity coefficient is a standard metric for evaluating image segmentation [27] and measures the similarity between logical image masks (ground truth and test image). Similar approaches have been used to evaluate the performance of interpretability methods in previous studies [5, 45]. In this experiment, the ground truth is obtained after binarizing the Grad-CAM interpretation map and the test image corresponds to the binarization resulting from the interpretation method to be evaluated. The binarization method used was thresholding with threshold value set at 127. We use only strategy-1 for this experiment. Table 2 shows that LSFT-GAN has the highest average dice coefficient (0.6588).

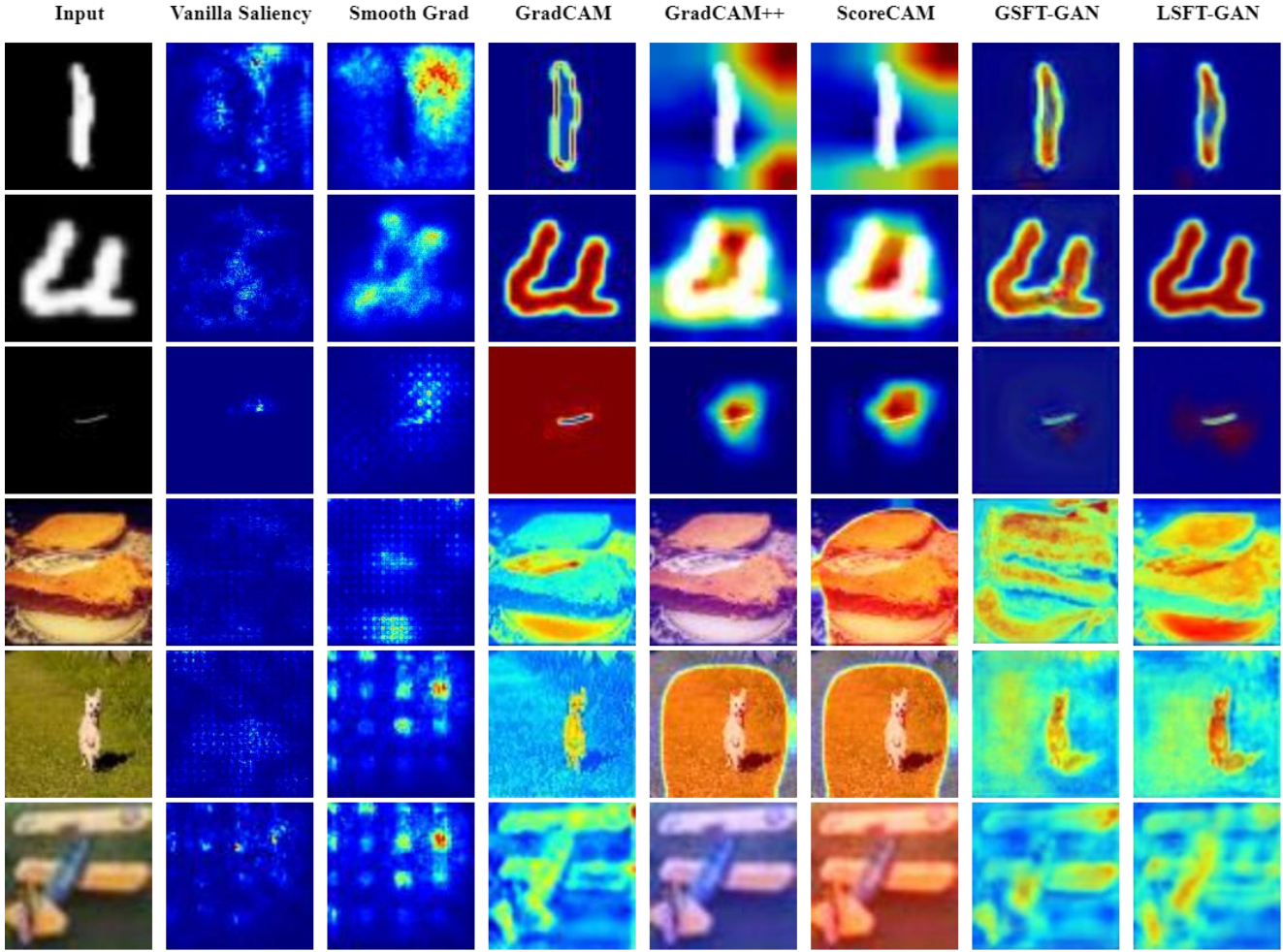

**Figure 5:** Visualization results of Vanilla Saliency [37], SmoothGrad [40],Grad-CAM [34],Grad-CAM++ [5], Scorecam [45]and our proposed SFT-GANs - GSFT-GAN and LSFT-GAN

### 4.2.3. Faithfulness Evaluation using Object Recognition

We evaluate the faithfulness of the interpretation maps generated by SFT-GANs using a slightly modified version of the metric proposed by Aditya et al.[5]. Point-wise multiplication of the input image and the interpretation map was computed, denoted as the energy map. An energy map highlights the key areas identified by the interpretation map. Object recognition was performed on the energy map, and the object recognizer's confidence exposes SFT-GAN's ability to produce faithful interpretations. The confidence here represents the likelihood of the object recognizer predicting the correct class. Three different metrics were used to evaluate faithfulness: (i) average drop %, (ii) increase in confidence, and (iii) win %. Although energy maps highlight essential regions, confidence is expected to drop. A lower drop would indicate that the interpretations are faithful. Average drop % is used to evaluate the drop in confidence, which is defined as

$$\text{.}(\%) = \frac{\sum E - O}{=} * 100 \qquad (7)$$

where is the confidence of the original image and is the confidence of the energy map. We ignored the cases in which the confidence of energy maps was greater than in the original images, as opposed to [5]. The increase in confidence metric was modified to gauge these cases. The average increase in confidence is defined as:

$$\text{.}(\%) = \frac{\sum O_i - E_i}{=} * 100 \qquad (8)$$

Win ratio is defined as the number of occurrences in which confidence of is greater than where and are energy maps of two interpretability methods. Table 3 displays the results of the experiment. LSFT-GAN outperformed both GSFT-GAN and Grad-CAM in terms of all three metrics. When compared to GSFT-GAN and Grad-CAM, LSFT-GAN has a higher percentage of increase in confidence (6.82%) and win (53.9%) and a lower average drop(%) (9.95%). Grad-CAM outperformed GSFT-GAN in terms of average drop % and win %, which shows the advantage of using LSFT-GAN over GSFT-GAN. Faithfulness is

| Metric | Grad-CAM | GSFT | LSFT |
|---|---|---|---|
| Avg Drop % | 11.49 | 11.73 | 9.95 |
| Avg Increase % | 6.57 | 6.10 | 6.82 |
| Win % | – | 49.0 | 53.9 |

**Table 3**
Results of Objective Evaluation. A lower average drop is better, and higher average increase is better. Win percentage is calculated compared with Grad-CAM, a higher win percentage is better. The metrics whose value is better when higher is marked in RED and better when lower is marked in BLUE further evaluated in Section 4.4 by degrading the input image in the presence of noise.

### 4.3. Qualitative Assessment
We show the visual comparision of proposed SFT-GANs verses state-of-the-art CNN explainability models in Section 4.3.1. Both of the SFT-GANs were able to identify additional regions of importance, although trained only with Grad-CAMs as target images, as shown in Section 4.3.2. We show that training using multiple datasets is effective in Section 4.3.3. We assess the performance of LSFT-GAN on unseen datasets and architectures in Section 4.3.4. We compare the proposed training strategies in Section 4.3.5. We display results of using LSFT-GAN to interpret an overfitted CNN in Section 4.3.6. We show the results of applying LSFT-GAN to image captioning in Section 4.3.7.

#### 4.3.1. Visual Comparison
In this section, we compare the interpretation maps produced by the SFT-GANs with state-of-the-art interpretability methods [37, 40, 34, 5, 45]. The first three rows of Figure 5 display interpretations for an image randomly selected from each of MNIST [11], EMINST [9] and Chinese [29] datasets that are used in the comparison. These datasets were used to train the GAN using strategy-1 (similar semantics). Rows 4,5,6 contain images from Food-11, Animals-10, and CIFAR-10 datasets respectively. These SFT-GAN results shown for these images are interpreted using SFT-GANs trained using strategy-2.

#### 4.3.2. Analogy with Vanilla Saliency
Although the proposed SFT-GANs were conditioned on Grad-CAM-based priors (CGMAs), the models were able to locate additional regions of significance, as shown in Figure 6. To validate this further, we compare the generated interpretation map to the interpretations generated by vanilla saliency [37] for the same dataset-model pair. As it can be seen in Figure 6, the additional relevant regions captured by SFT-GANs are analogous to that of Vanilla Saliency. Here we show only the results on the MNIST dataset interpreted using LSFT-GAN trained with strategy-1.

#### 4.3.3. Advantage of using multiple Datasets
In this section, we explore training the GAN using CG-MAs from 3 different datasets under each training strategy. As shown in Figure 7, the LSFT-GAN generalized better

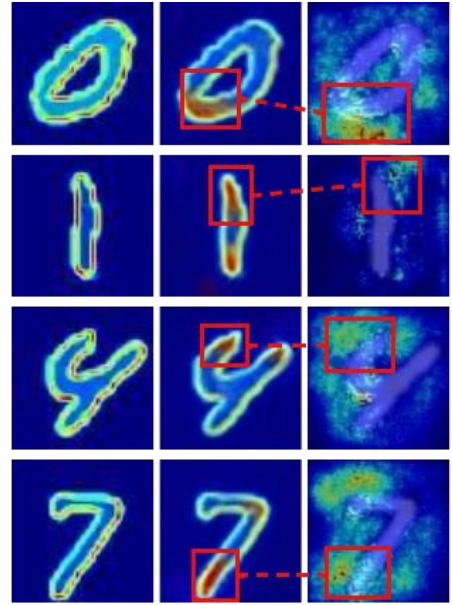

**Figure 6:** Analogies between LSFT and Vanilla Saliency on interpreting how VGG16 classifies MNIST [11]. Column 1 represents Grad-CAMs column 2 represents LSFT and Column 3 represents Vanilla Saliency. Red squares are used to highlight analogous regions

when provided with data from 3 datasets. The usage of multiple datasets aided the SFT-GANs to understand CNNs faster. While the LSFT-GAN trained using a single dataset was able to generate meaningful interpretations after 6 epochs, the LSFT-GAN trained with 3 datasets required only a single epoch to produce meaningful interpretations. Since the LSFT-GAN started to produce meaningful interpretations from the first epoch, we can infer that the LSFT-GAN generalized to CNNs while training with multiple datasets.

#### 4.3.4. Interpretations on unseen datasets and models
In this section, we evaluate the performance of LSFT-GAN on novel datasets and architectures unseen during the training. For this experiment, we used LSFT-GAN trained using strategy-2 and the datasets the Cats Vs Dogs [13], EMINST [9] and 30 Musical Instruments[15]. An Inception V3[44] was used to classify Cats vs Dogs dataset, EMINST was classified using Xception[8] and Musical Instruments were classified using VGG-19 [38]. Although EMINST was used to train LSFT-GAN in strategy-1 it is unseen in strategy-2. Figure 8 shows the results obtained by the LSFT-GAN. The Cats Vs Dog dataset classified with an Inception V3 achieved an accuracy of over 99% on the test set. Hence, the highly relevant regions should be regions of the object. This was reflected in the interpretation maps generated by LSFT-GAN. The Xception model trained on EMINST dataset achieved a test accuracy of 97 %. Although both LSFT-GAN and Grad-CAM highlighted complete alphabets, LSFT-GAN managed to highlight important regions with more intensity. For the instruments dataset both Grad-CAM and LSFT-GAN managed to highlight important

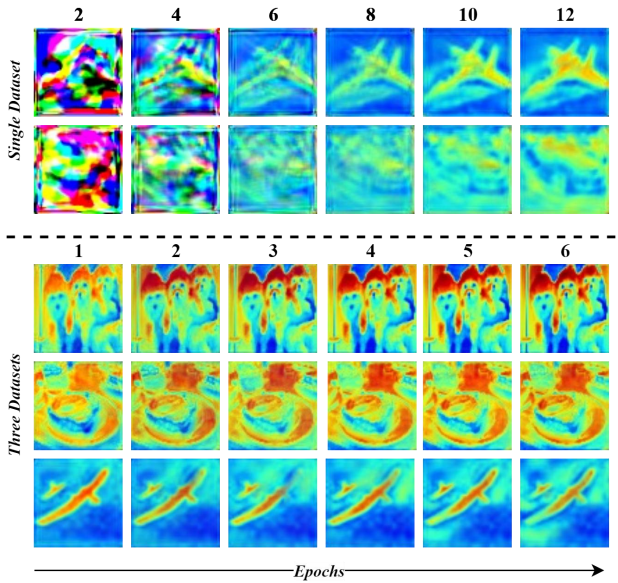

**Figure 7:** Epoch-wise interpretation maps generated by LSFT-GAN trained on a single dataset (CIFAR 10[22]) and LSFT-GAN trained on three datasets(CIFAR-10[22], Animals-10[3], Food-11[39]). The interpretation maps shown in the figure are generated from the testing set of respective datasets

regions, but Grad-CAM was better than LSFT-GAN. We believe the reason Grad-CAM outperformed LSFT-GAN because LSFT-GAN was trained on datasets with relatively different semantics. Large-scale training of LSFT-GAN could alleviate this issue. From these results, we could conclude that LSFT-GAN trained using strategy-2 does indeed generalize to unseen datasets and architectures.

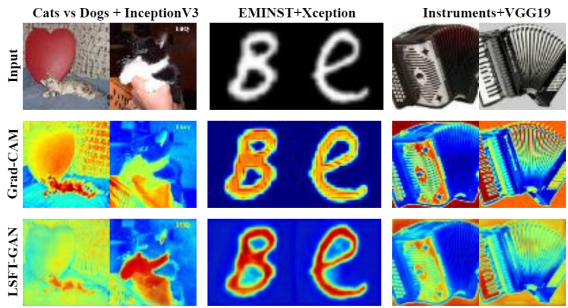

**Figure 8:** Interpretation Maps generated using LSFT-GAN for Unseen Dataset-Architecture combinations

### 4.3.5. Comparing Training Strategies

In this section, we compare the results of LSFT-GAN trained using the two training strategies for interpreting unseen datasets. InceptionV3 model trained on Cats Vs Dog dataset and VGG16 model trained on Tamil Characters dataset [35] were used for comparison. We chose the datasets to identify which among the strategies is better for datasets with similar semantics and dissimilar semantics. Figure 9 shows the results obtained. For the cats vs dog dataset, both datasets were able to localize regions of interest, but LSFT-GAN trained using strategy-1 is found to identify relevant intensity for the identified regions of interest. Through this, we found that the LSFT-GAN generated using strategy-2 is more general than the interpretations generated by the LSFT-GAN trained using strategy-1. For the Tamil Characters dataset, both strategies were able to identify regions of interest. On comparing the regions of interest generated by both strategies, we found the interpretations generated by strategy-1 to identify specific regions of interest, which could come in handy while interpreting highly critical systems.

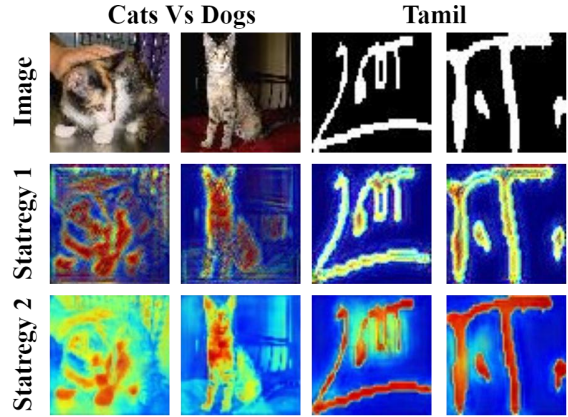

**Figure 9:** Comparing the performance of the training strategies against unseen datasets

### 4.3.6. Interpreting an Overfitted CNN

So far, the experiments were only applied on well-trained CNNs i.e. in the absence of overfitting. In this section, we compare which among LSFT-GAN and Grad-CAM is better for an overfitted CNN model. The dataset used for this experiment is the People Playing Musical Instruments Dataset[49], which consists of images of people holding and playing musical instruments. Each image is assigned to 12 classes according to the musical instruments present. Due to the presence of unique features such as clothes and background, the CNN models tend to memorize these features during training, which leads to overfitting. We trained an Inception V3 architecture to classify the People Playing Musical Instruments dataset. Figure 10 shows the LSFT and Grad-CAM interpretations for the overfitted model. We can find that both LSFT-GAN and Grad-CAM identified background regions as important. Mainly, clothes and background artifacts were highlighted. Our LSFT-GAN was able to identify additional regions of importance when compared to Grad-CAM.

### 4.3.7. Interpreting an Image Captioning Model

We experimented LSFT-GAN's ability to interpret CNNs used for an image captioning task. The image captioning model was trained on a subset of 6000 images on MS

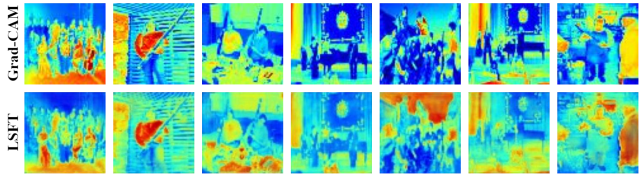

**Figure 10:** Maps generated for interpreting People Playing Musical Instruments dataset trained using an overfitted Inception V3 architecture. The results show that the CNN focuses on clothes and background objects to make its predictions rather than musical instruments.

COCO dataset[26]. We trained the dataset using a TensorFlow repository.[1]. Figure 11 shows the interpretation results. LSFT-GAN was able to reasonably interpret image captioning. When compared with Grad-CAM, it was able to localize subjects better. The verbs were given a relatively lesser intensity than subjects when interpreted using LSFT-GAN for example, in the third row of Figure 11.

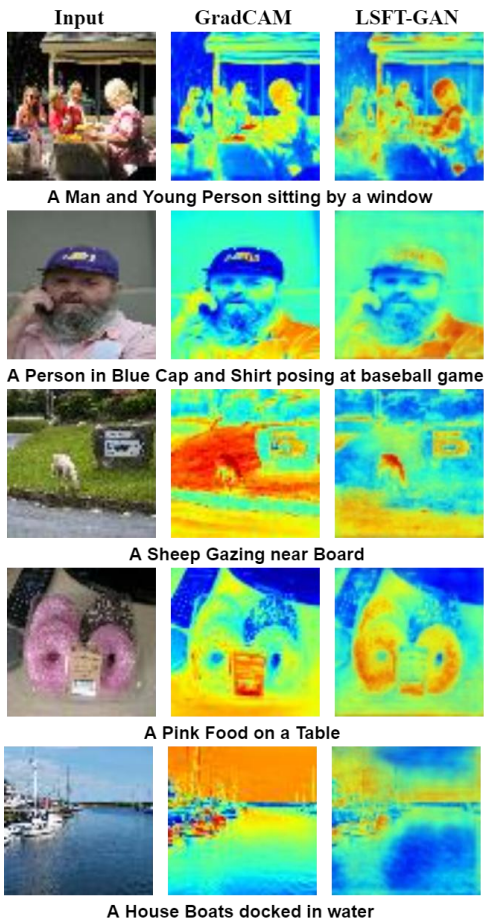

**Figure 11:** Results of applying Grad-CAM and LSFT-GAN on an image captioning task

---

[1]TensorFlow Image Captioning Repository

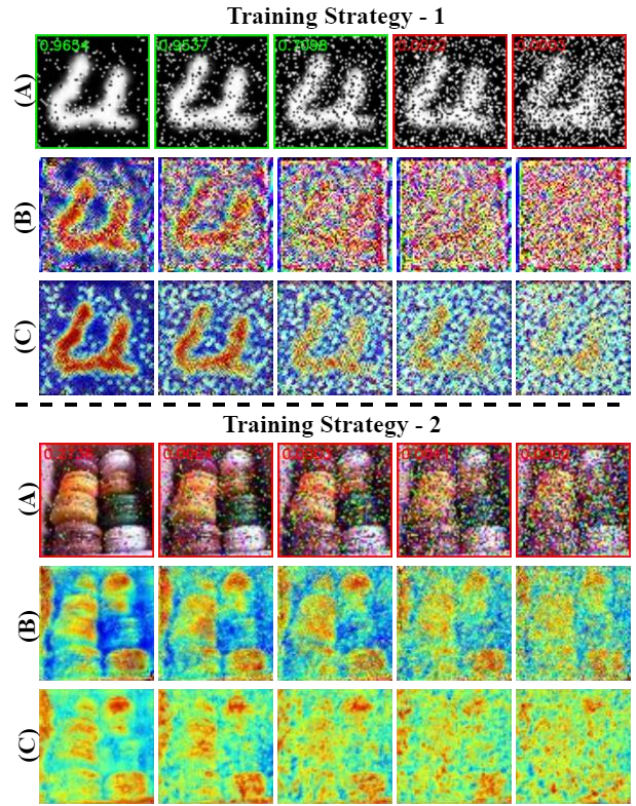

**Figure 12:** Effect on interpretation maps on degrading the input images with an increasing amount of salt and pepper noise. The amount of noise added varies from 10 % to 50% with an increase of 10% per image. Row (A) contains noise-added inputs used for classification. Row (B) displays GSFT interpretations. Row (C) displays LSFT interpretations.

### 4.4. Robustness towards image degradation

**Robustness to noise degradation** An interpretability method can be trusted if it is resistant to noisy degradation of input images. We degraded the input images by the incremental addition of salt-and-pepper noise. The degraded input images are then classified using CNNs. CGMAs were obtained and passed into SFT-GANs to produce interpretations of the classification. Salt-and-pepper noise was used to deter the input images by randomly setting the pixels of the input image to either 0 or 255. The degree of degradation (from 0.05 to 0.5. in this experiment) is the fraction of pixels affected by salt-and-pepper noise. We performed the test on both training strategies.

In Figure 12 for strategy-1, GSFT-GAN misinterpreted noise as key image regions (red regions in heat map (Figure 12.b). The noisy regions were colored as bluish-green by the LSFT-GAN, implying that the LSFT-GAN was able to distinguish between the noisy and image-specific pixels. The gradual increase in noise in the input images validated the resistance of LSFT-GAN to degradation. When the amount of noise was increased, the output map from LSFT-GAN remained consistent in identifying image-specific regions rather than the noisy pixels. The key image-specific regions vanished faster in GSFT-GANs than in LSFT-GANs in terms of the amount of noise added.

**Trustfulness Evaluation** Adebayo *et al.*[1] assesses the quality of interpretability methods by randomizing model parameters. Inspired by this method, we quantified the maps acquired by applying SFT-GANs to interpret noisy images. As shown in Figure 12 on the addition of noise, the background is misinterpreted as an important region in GSFT-GAN. Thus, we use the average background error as a metric to evaluate the trustfulness, as shown in Equation 9. We compute the average background error by gradually introducing noise into the image. As shown in Figure 13, LSFT-GAN produces a lower average background error than GSFT-GAN, and the gap between the two SFT-GANs gradually widens; thus, interpretations using LSFT-GAN can be trusted more than those using GSFT-GAN.

$$. = \frac{\sum_{=1} (1 - GT) * M}{} \quad (9)$$

Where GT is the ground truth of an input image, (1-GT) is the ground truth background of the image and IM is the interpretation map produced by an SFT-GAN.

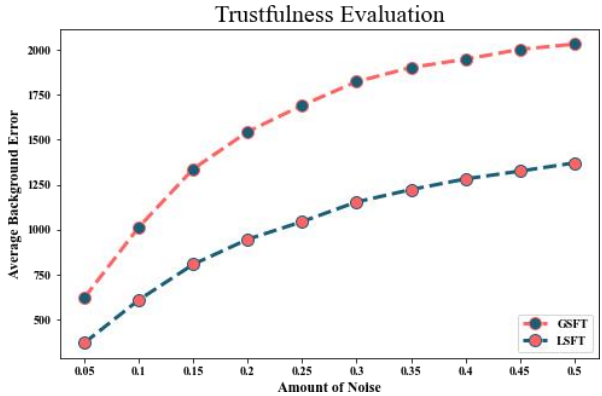

**Figure 13:** Trustfulness evaluation plot. The graph plots average background error therefore lower value is better

## 5. Interpreting LSFT-GAN

While our LSFT-GAN was able to generate meaningful interpretations better than SOA interpretation methods, it being a neural network is not transparent. It is important to understand how the LSFT-GAN interprets the CGMAs to understand CNNs. We thus introduce a method to interpret LSFT-GAN in this section. We also analyze the results to present interpretations on the working of LSFT-GAN.

### 5.1. LSFT-GAN Interpretation Method

Via this interpretation method we intend to identify the importance of each input CGMA to make predictions. To indentify the importance of a CGMA $C$ from a set of CGMAs $S$, we made the CGMA $C = 0$ of the resulting set of CGMAs( $S^*$) is passed into the LSFT-GAN $L$ to produce $L(S^*)$. We then calculate the relevance of that CGMA using Equation 10. The relevance, in words, is the total difference between $L(S)$ and $L(S^*)$. Since the relevance is calculated CGMA-wise we would be able to witness the regions of the input CNN that are more important for interpreting the CNN.

$$Relevance = \sum_{=1} \sum_{=1} |L(S)[i, j] - L(S^*)[i, j]| \quad (10)$$

where m,n are the dimension of the interpretation map generated.

### 5.2. Interpreting the LSFT-GAN Results for Individual image-architecture pairs

We show the results of interpreting the LSFT-GAN on three dataset-architecture pairs. EMIST-Xception, CatsVsDogs-InceptionV3 and Musical Instruments-VGG19. These datasets and architectures were unseen during the training of LSFT-GAN. Figure 15 shows the results of interpreting the LSFT-GAN. Taking the image from row 2 column 1 as an example, we can find that the removal of CGMA 3 does not cause much change to the final interpretation map and that CGMA 7 contributes the most to the interpretation map. For the interpretation map present in row 3 and column 3 all CGMAs were key to construct the final interpretation map.

### 5.3. Interpreting the average behavior of LSFT-GAN towards dataset-architecture pairs

We calculated the average relevance of each CGMA for interpreting dataset-architecture pairs. Figure 16 shows the visual difference witnessed in removing CGMAs. The average relevance of each CGMA for each dataset-architecture pair is shown in Figure 15. From the graphs, we can infer that the initial layers of the CNN are equally important as the final layers for interpretation. This inference suggests that the approximate interpretations generated by gradient-based methods from the final layers might not be as accurate as methods that interpret based on the complete model, such as our LSFT-GAN. From Figure 16 we could notice two main inferences. First, the sensitivity towards the intensity of importance is lessened on removing the last CGMA. Second, the removal of initial CGMAs leads to poor localization of important regions. From these inferences, we can conclude that the LSFT-GAN looks into the initial layers to identify regions of importance while the CGMAs towards the end are responsible to adjust the importance of each identified region.

## 6. Human Comprehensibility

In this section, we compare our proposed LSFT-GAN and Grad-CAM to identify the better CNN interpretation method. We conducted two surveys to identify the better interpretation method for classification and image captioning tasks.

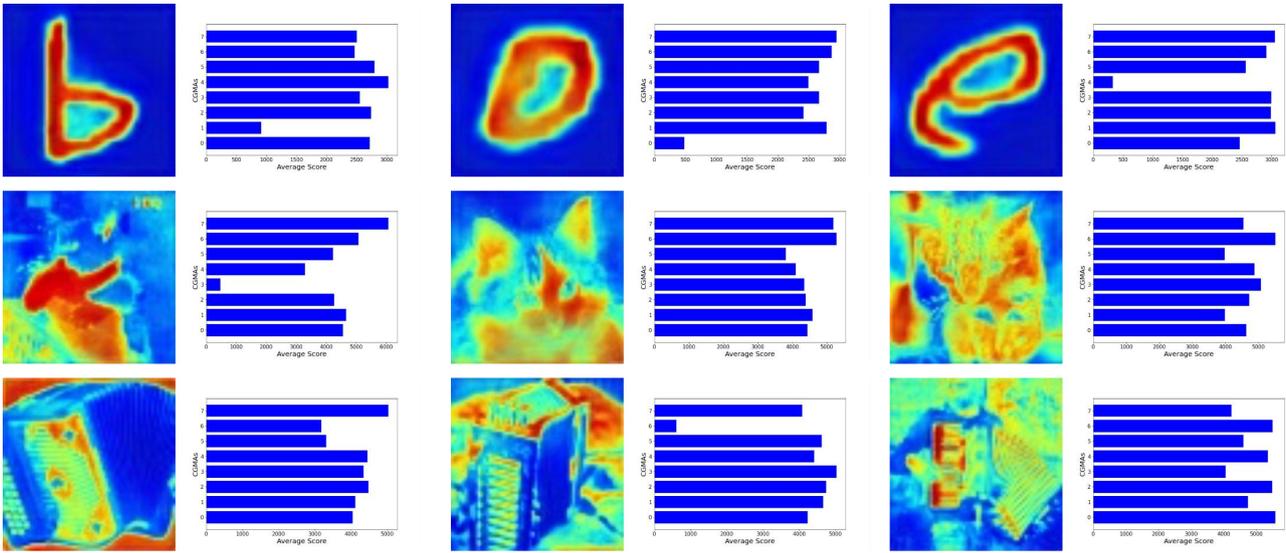

**Figure 14:** Interpretations on why LSFT-GAN produced specific interpretation maps. Each graph's Y axis represents a CGMA and X axis represent the relevance of the CGMA. Zoom in to look into the graphs clearly

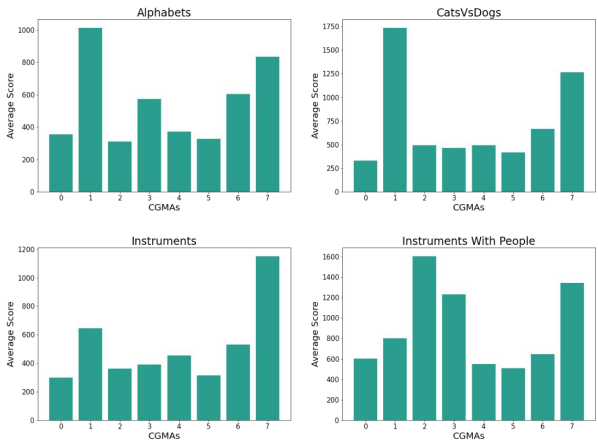

**Figure 15:** The average relevance of each CGMA for each dataset. The X-axis represents CGMAs and the Y-axis represents the relevance of the CGMA

### 6.1. Image Classification Survey

The survey contained 20 questions. Each question consisted of an image and its corresponding LSFT-GAN and Grad-CAM Interpretations. 25 volunteers took the survey. Out of the 20 questions, the volunteers felt LSFT-GAN provided better interpretations for 16 questions. The volunteers felt that Grad-CAM was better than LSFT-GAN for 3 questions and LSFT-GAN and Grad-CAM was deemed to produce equally understandable interpretations for 1 question. From this survey, we can derive that the volunteers felt LSFT-GAN provides better interpretations for the majority of the questions. The results of the survey are displayed in Figure 17

### 6.2. Image Captioning Survey

The survey contained 10 questions. Each question had an image, the generated caption, and their LSFT and Grad-CAM interpretations. This survey was attended by 25 volunteers. The volunteers found the interpretations provided by LSFT-GAN better than Grad-CAM for 7 questions and the interpretations provided by Grad-CAM to be better for 3 questions. From this survey, we can understand that the volunteers felt that LSFT-GAN interpretations were better than Grad-CAM interpretations despite LSFT-GAN used being trained on image classification tasks. The results of the survey are displayed in Figure 18.

## 7. Limitations

SFT-GANs are successful in interpreting CNNs. The SFT-GANs provide better interpretations than existing interpretation methods but, the computational cost for the interpretation generation using SFT-GANs is higher compared to interpretation methods such as Grad-CAM[34] and Grad-CAM++[5].

## 8. Conclusion

With the objective of interpreting CNNs using GANs, we introduced a method to represent the CNN as a condition of the GAN. We introduced two architectures, GSFT-GAN and LSFT-GAN, providing different conditioning strategies. We evaluated the performance both qualitatively and quantitatively using 3 different CNN architectures and 6 different datasets. Through the experiments, we show that the interpretations generated by the SFT-GANs are analogous to the state-of-the-art models. We found that locally conditioning the GAN is better than globally conditioning the GAN. LSFT-GAN outperformed Grad-CAM both visually and quantitatively. We found that our LSFT-GAN was able to interpret unseen CNN architectures trained on unseen datasets. We also found that our LSFT-GAN was able to interpret underfitting CNN models. Our LSFT-GAN was found to be task-agnostic. We have explored different training strategies and found that LSFT-GAN was able to provide general interpretations and that LSFT-GAN could be trained for specific tasks to get more specific results. The method is robust to the presence of image degradation and allows better generalization by using multiple datasets to generate visual interpretations of a CNN architecture. On interpreting our LSFT-GAN we found out that it used the initial CNN layers and final CNN layers predominantly for interpreting CNNs. We conclude that conditional GAN constitute a suitable and promising approach to generating visual interpretability maps of CNN models with potential applications in the field of deep learning interpretability.

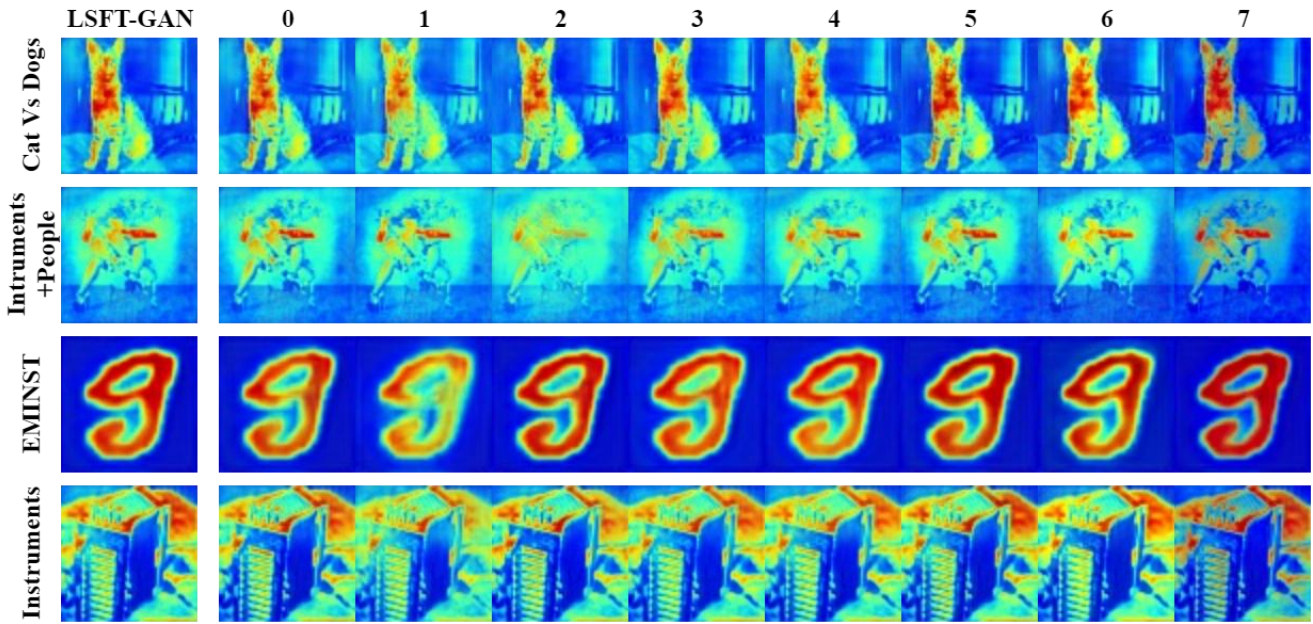

**Figure 16:** Interpretation maps generated by LSFT-GAN after the blackout of a specific CGMA. There are a total of 8 CGMAs (0-7)

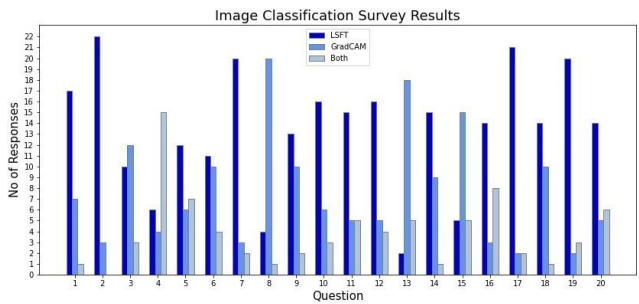

**Figure 17:** Results of image classification human comprehension survey

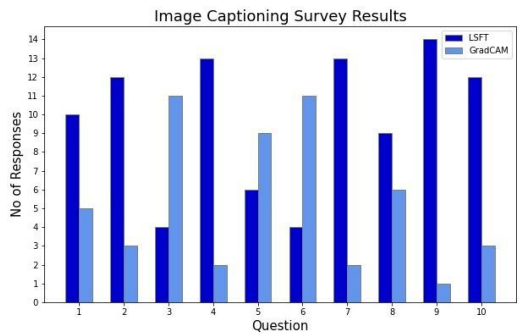

**Figure 18:** Results of human comprehension survey based on image captioning interpretations

## 9. Acknowledgements

This research was funded by the Spanish Ministry of Science and Innovation, grant number PID2020-116927RB-C22 (R.B.).

### CRediT authorship contribution statement

**R T Akash Guna:** Data curation, Investigation, Software, Validation, Visualization, Writing – original draft. **Raul Benitez:** Conceptualization, Funding acquisition, Investigation, Methodology, Project administration, Supervision, Writing – review & editing. **O K Sikha:** Conceptualization, Investigation, Methodology, Validation, Writing – review & editing.

## References

[1] Adebayo, J., Gilmer, J., Muelly, M., Goodfellow, I., Hardt, M., Kim, B., 2018. Sanity checks for saliency maps. Advances in neural information processing systems 31.

[2] Aishwarya, G., Krishnan, K.R., 2021. Generative adversarial networks for facial image inpainting and super-resolution, in: Journal of Physics: Conference Series, IOP Publishing. p. 012103.

[3] Alessio, C., 2019. Animals-10. URL: https://www.kaggle.com/datasets/alessiocorrado99/animals10.

[4] Bengio, Y., Simard, P., Frasconi, P., 1994. Learning long-term dependencies with gradient descent is difficult. IEEE Transactions on Neural Networks 5, 157–166. doi:10.1109/72.279181.

[5] Chattopadhay, A., Sarkar, A., Howlader, P., Balasubramanian, V.N., 2018. Grad-cam++: Generalized gradient-based visual explanations for deep convolutional networks, in: 2018 IEEE winter conference on applications of computer vision (WACV), pp. 839–847.

[6] Cherian, A., Sullivan, A., 2019. Sem-gan: semantically-consistent image-to-image translation, in: 2019 IEEE winter conference on applications of computer vision (wacv), IEEE. pp. 1797–1806.

[7] Choi, Y., Choi, M., Kim, M., Ha, J.W., Kim, S., Choo, J., 2018. Stargan: Unified generative adversarial networks for multi-domain image-to-image translation, in: Proceedings of the IEEE conference on computer vision and pattern recognition, pp. 8789–8797.

[8] Chollet, F., 2017. Xception: Deep learning with depthwise separable convolutions, in: Proceedings of the IEEE conference on computer vision and pattern recognition, pp. 1251–1258.

[9] Cohen, G., Afshar, S., Tapson, J., van Schaik, A., 2017. EMNIST: an extension of MNIST to handwritten letters. CoRR abs/1702.05373. URL: http://arxiv.org/abs/1702.05373, arXiv:1702.05373.

[10] Deng, J., Dong, W., Socher, R., Li, L.J., Li, K., Fei-Fei, L., 2009. Imagenet: A large-scale hierarchical image database, in: 2009 IEEE Conference on Computer Vision and Pattern Recognition, pp. 248–255. doi:10.1109/CVPR.2009.5206848.

[11] Deng, L., 2012. The mnist database of handwritten digit images for machine learning research. IEEE Signal Processing Magazine 29, 141–142.

[12] Doran, D., Schulz, S., Besold, T.R., 2017. What does explainable ai really mean? a new conceptualization of perspectives. arXiv preprint arXiv:1710.00794 .

[13] Elson, J., Douceur, J.R., Howell, J., Saul, J., 2007. Asirra: a captcha that exploits interest-aligned manual image categorization. CCS 7, 366–374.

[14] Ge, Y., Song, Y., Zhang, R., Ge, C., Liu, W., Luo, P., 2021. Parser-free virtual try-on via distilling appearance flows, in: Proceedings of the IEEE/CVF Conference on Computer Vision and Pattern Recognition, pp. 8485–8493.

[15] Gerry, 2021. 30 musical instruments -image classification. URL: https://www.kaggle.com/datasets/gpiosenka/musical-instruments-image-classification.

[16] Goodfellow, I., Pouget-Abadie, J., Mirza, M., Xu, B., Warde-Farley, D., Ozair, S., Courville, A., Bengio, Y., 2014. Generative adversarial nets. Advances in neural information processing systems 27.

[17] He, K., Gkioxari, G., Dollár, P., Girshick, R., 2017. Mask r-cnn, in: Proceedings of the IEEE international conference on computer vision, pp. 2961–2969.

[18] He, K., Zhang, X., Ren, S., Sun, J., 2016. Deep residual learning for image recognition, in: Proceedings of the IEEE conference on computer vision and pattern recognition, pp. 770–778.

[19] Honda, S., 2019. Viton-gan: Virtual try-on image generator trained with adversarial loss. arXiv preprint arXiv:1911.07926 .

[20] Howard, A.G., Zhu, M., Chen, B., Kalenichenko, D., Wang, W., Weyand, T., Andreetto, M., Adam, H., 2017. Mobilenets: Efficient convolutional neural networks for mobile vision applications. arXiv preprint arXiv:1704.04861 .

[21] Isola, P., Zhu, J.Y., Zhou, T., Efros, A.A., 2017. Image-to-image translation with conditional adversarial networks, in: Proceedings of the IEEE conference on computer vision and pattern recognition, pp. 1125–1134.

[22] Krizhevsky, A., 2012. Learning multiple layers of features from tiny images. University of Toronto URL: https://api.semanticscholar.org/CorpusID:18268744.

[23] Lata, K., Dave, M., Nishanth, K., 2019. Image-to-image translation using generative adversarial network, in: 2019 3rd International conference on Electronics, Communication and Aerospace Technology (ICECA), IEEE. pp. 186–189.

[24] LeCun, Y., Bengio, Y., et al., . Convolutional networks for images, speech, and time series. The Handbook of Brain Theory and Neural Networks .

[25] Ledig, C., Theis, L., Huszár, F., Caballero, J., Cunningham, A., Acosta, A., Aitken, A., Tejani, A., Totz, J., Wang, Z., et al., 2017. Photo-realistic single image super-resolution using a generative adversarial network, in: Proceedings of the IEEE conference on computer vision and pattern recognition, pp. 4681–4690.

[26] Lin, T.Y., Maire, M., Belongie, S., Hays, J., Perona, P., Ramanan, D., Dollár, P., Zitnick, C.L., 2014. Microsoft coco: Common objects in context, in: European conference on computer vision, Springer. pp. 740–755.

[27] Milletari, F., Navab, N., Ahmadi, S.A., 2016. V-net: Fully convolutional neural networks for volumetric medical image segmentation, in: 2016 fourth international conference on 3D vision (3DV), IEEE. pp. 565–571.

[28] Mirza, M., Osindero, S., 2014. Conditional generative adversarial nets. arXiv preprint arXiv:1411.1784 .

[29] Nazarpour, K., Chen, M., 2017. Handwritten chinese numbers. URL: https://data.ncl.ac.uk/articles/dataset/Handwritten_Chinese_Numbers/10280831/1, doi:10.17634/137930-3.

[30] Patil, S.O., Sajith Variyar., V., Soman, K.P., 2020. Speed bump segmentation an application of conditional generative adversarial network for self-driving vehicles, in: 2020 Fourth International Conference on Computing Methodologies and Communication (ICCMC), pp. 935–939. doi:10.1109/ICCMC48092.2020.ICCMC-000173.

[31] Ronneberger, O., Fischer, P., Brox, T., 2015. U-net: Convolutional networks for biomedical image segmentation, in: International Conference on Medical image computing and computer-assisted intervention, Springer. pp. 234–241.

[32] Sanjay, A., Nair, J.J., Gopakumar, G., 2021. Haze removal using generative adversarial network. Advances in Computing and Network Communications , 207–217.

[33] Schroff, F., Kalenichenko, D., Philbin, J., 2015. Facenet: A unified embedding for face recognition

and clustering, in: 2015 IEEE Conference on Computer Vision and Pattern Recognition (CVPR), pp. 815–823. doi:10.1109/CVPR.2015.7298682.

[34] Selvaraju, R.R., Cogswell, M., Das, A., Vedantam, R., Parikh, D., Batra, D., 2017. Grad-cam: Visual explanations from deep networks via gradient-based localization, in: ICCV, pp. 618–626. doi:10.1109/ICCV.2017.74.

[35] Shaffi, N., Hajamohideen, F., 2021. uthcd: A new benchmarking for tamil handwritten OCR. CoRR abs/2103.07676. URL: https://arxiv.org/abs/2103.07676, arXiv:2103.07676.

[36] Sikha, O., Bharath, B., 2022. Vgg16-random fourier hybrid model for masked face recognition. Soft Computing , 1–16.

[37] Simonyan, K., Vedaldi, A., Zisserman, A., 2013. Deep inside convolutional networks: Visualising image classification models and saliency maps. arXiv preprint arXiv:1312.6034 .

[38] Simonyan, K., Zisserman, A., 2014. Very deep convolutional networks for large-scale image recognition. arXiv preprint arXiv:1409.1556 .

[39] Singla, A., Yuan, L., Ebrahimi, T., 2016. Food/non-food image classification and food categorization using pre-trained googlenet model, in: Proceedings of the 2nd International Workshop on Multimedia Assisted Dietary Management, pp. 3–11.

[40] Smilkov, D., Thorat, N., Kim, B., Viégas, F., Wattenberg, M., 2017. Smoothgrad: removing noise by adding noise. arXiv preprint arXiv:1706.03825 .

[41] Sorensen, T.A., 1948. A method of establishing groups of equal amplitude in plant sociology based on similarity of species content and its application to analyses of the vegetation on danish commons. Biol. Skar. 5, 1–34.

[42] Springenberg, J.T., Dosovitskiy, A., Brox, T., Riedmiller, M., 2014. Striving for simplicity: The all convolutional net. arXiv preprint arXiv:1412.6806 .

[43] Srihari, K., Sikha, O., 2022. Partially supervised image captioning model for urban road views, in: Intelligent Data Communication Technologies and Internet of Things. Springer, pp. 59–73.

[44] Szegedy, C., Vanhoucke, V., Ioffe, S., Shlens, J., Wojna, Z., 2016. Rethinking the inception architecture for computer vision, in: Proceedings of the IEEE conference on computer vision and pattern recognition, pp. 2818–2826.

[45] Wang, H., Wang, Z., Du, M., Yang, F., Zhang, Z., Ding, S., Mardziel, P., Hu, X., 2020. Score-cam: Score-weighted visual explanations for convolutional neural networks, in: Proceedings of the IEEE/CVF conference on computer vision and pattern recognition workshops, pp. 24–25.

[46] Wang, X., Li, Y., Zhang, H., Shan, Y., 2021. Towards real-world blind face restoration with generative facial prior, in: Proceedings of the IEEE/CVF conference on computer vision and pattern recognition, pp. 9168–9178.

[47] Wang, X., Yu, K., Dong, C., Loy, C.C., 2018. Recovering realistic texture in image super-resolution by deep spatial feature transform, in: Proceedings of the IEEE conference on computer vision and pattern recognition, pp. 606–615.

[48] Wu, H., Zheng, S., Zhang, J., Huang, K., 2019. Gp-gan: Towards realistic high-resolution image blending, in: Proceedings of the 27th ACM international conference on multimedia, pp. 2487–2495.

[49] Yao, B., Fei-Fei, L., 2010. Grouplet: A structured image representation for recognizing human and object interactions, in: 2010 IEEE Computer Society Conference on Computer Vision and Pattern Recognition, IEEE. pp. 9–16.

[50] Yu, X., Qu, Y., Hong, M., 2018. Underwater-gan: Underwater image restoration via conditional generative adversarial network, in: International Conference on Pattern Recognition, Springer. pp. 66–75.

[51] Zeiler, M.D., Fergus, R., 2014. Visualizing and understanding convolutional networks, in: European conference on computer vision, Springer. pp. 818–833.

[52] Zhou, B., Khosla, A., Lapedriza, A., Oliva, A., Torralba, A., 2016. Learning deep features for discriminative localization, in: Proceedings of the IEEE conference on computer vision and pattern recognition, pp. 2921–2929.

[53] Zhu, J.Y., Park, T., Isola, P., Efros, A.A., 2017. Unpaired image-to-image translation using cycle-consistent adversarial networks, in: Proceedings of the IEEE international conference on computer vision, pp. 2223–2232.